\documentclass{article}

\usepackage[preprint]{neurips_2025}

\usepackage[utf8]{inputenc} % allow utf-8 input
\usepackage[T1]{fontenc}    % use 8-bit T1 fonts
\usepackage{hyperref}       % hyperlinks
\usepackage{url}            % simple URL typesetting
\usepackage{booktabs}       % professional-quality tables
\usepackage{amsfonts}       % blackboard math symbols
\usepackage{nicefrac}       % compact symbols for 1/2, etc.
\usepackage{microtype}      % microtypography
\usepackage{xcolor}         % colors
\usepackage{graphicx}

\usepackage{multirow}
\usepackage{makecell}
\usepackage{tabularx}
\usepackage{wrapfig}
\usepackage{caption}
\usepackage{enumitem}

\title{BrowseComp-ZH: Benchmarking Web Browsing Ability of Large Language Models in Chinese}

\author{%
  Peilin Zhou$^{1\dagger}$\thanks{Corresponding \quad
  $^{\dagger}$Co-first authors} \quad
  Bruce Leon$^{2\dagger}$ \quad
  Xiang Ying$^{3}$ \quad
  Can Zhang$^{2}$ \quad
  Yifan Shao \quad
  Qichen Ye$^{4}$ \\
  \textbf{Dading Chong}$^{2}$ \quad
  \textbf{Zhiling Jin}$^{5}$ \quad
  \textbf{Chenxuan Xie}$^{6}$ \quad
  \textbf{Meng Cao}$^{7}$ \quad
  \textbf{Yuxin Gu}$^{8}$ \\
  \textbf{Sixin Hong}$^{2}$ \quad
  \textbf{Jing Ren}$^{2}$ \quad
  \textbf{Jian Chen}$^{1, 9}$ \quad
  \textbf{Chao Liu}$^{2}$ \quad
  \textbf{Yining Hua}$^{10}$ \vspace{3mm}\\
  $^{1}$Hong Kong University of Science and Technology (Guangzhou) \\
  $^{2}$Peking University \quad
  $^{3}$Mindverse AI \quad
  $^{4}$Alibaba Group \quad
  $^{5}$Zhejiang University \\ 
  $^{6}$Zhejiang University of Technology \quad
  $^{7}$MBZUAI \quad
  $^{8}$NIO \quad
  $^{9}$HSBC \\
  $^{10}$Harvard T.H. Chan School of Public Health \\
  \texttt{\{zhoupalin, yifashao209, hello.crabboss, mengcaopku\}@gmail.com} \\
  \texttt{yingxiang@mindverse.ai,  hongsixin1995hk\}@gmail.com}\\
    \texttt{\{zhangcan, yeeeqichen, 1601213984, 1901210484\}@pku.edu.cn}\\
  \texttt{22415038@zju.edu.cn, 
jchen524@connect.hkust-gz.edu.cn} \\
  \texttt{chaoliu@pku.org.cn, 
Yininghua@g.harvard.edu, aaron.gu1@nio.com}
}

\begin{document}

\maketitle
\setcounter{footnote}{0}

\begin{abstract}
As large language models (LLMs) evolve into tool-using agents, the ability to browse the web in real-time has become a critical yardstick for measuring their reasoning and retrieval competence. Existing benchmarks such as BrowseComp concentrate on English and overlook the linguistic, infrastructural, and censorship-related complexities of other major information ecosystems—most notably Chinese. 
To address this gap, we introduce \textbf{BrowseComp-ZH}, a high-difficulty benchmark purpose-built to comprehensively evaluate LLM agents on the Chinese web.
BrowseComp-ZH consists of 289 multi-hop questions spanning 11 diverse domains. Each question is reverse-engineered from a short, objective, and easily verifiable answer (e.g., a date, number, or proper noun). A two-stage quality control protocol is applied to strive for high question difficulty and answer uniqueness. %: 1) filtering questions easily retrievable via major search engines, and 2) applying a human-in-the-loop process to eliminate those with non-unique answers based on LLM-aided reasoning and manual verification. Solving these high-difficulty questions requires agents to comprehend multiple constraints, navigate across webpages, and synthesize evidence, setting a high bar for browsing capabilities in the Chinese web context.

We benchmark over 20 state-of-the-art language models and agentic search systems on our proposed BrowseComp-ZH. Despite their strong conversational and retrieval capabilities, most models struggle severely: a large number achieve accuracy rates below 10\%, and only a handful exceed 20\%. Even the best-performing system, OpenAI's DeepResearch, reaches just 42.9\%. These results demonstrate the considerable difficulty of BrowseComp-ZH, where success demands not only effective retrieval strategies, but also sophisticated reasoning and information reconciliation—capabilities that current models still struggle to master.

Our dataset, construction guidelines, and benchmark results have been publicly released at \url{https://github.com/PALIN2018/BrowseComp-ZH}.
\end{abstract}

\section{Introduction}

As large language models (LLMs) evolve from static knowledge repositories to dynamic agents capable of using external tools, tasks that involve web browsing have emerged as a critical lens through which to evaluate their real-world reasoning and information-seeking capabilities~\cite{xi2025rise}. By interacting with search engines and navigating live web content, LLMs can augment their internal knowledge with up-to-date external evidence, retrieve context-specific information, and perform multi-hop reasoning across heterogeneous sources. This browsing ability extends the temporal scope of LLMs while enabling them to tackle questions that lie beyond the reach of pretraining, such as time-sensitive facts or obscure entity relations that require targeted retrieval.

While an increasing number of studies~\cite{li2023web,fernandez2024search,fan2024survey,vu2023freshllms,lai2025webglm,he2024zero,lai2024autowebglm,xiong2024search} demonstrate that web browsing greatly improves LLM performance on downstream tasks, there remains a surprising lack of direct evaluation of browsing capabilities themselves—i.e., the ability of LLMs to effectively retrieve, filter, and reason over information from the web. This evaluation is crucial for assessing the true web-browsing competence of LLMs and understanding their potential to tackle real-world tasks that require dynamic information retrieval. To address this gap, \cite{wei2025browsecomp} introduced a benchmark of reverse-designed, evidence-grounded queries that challenge English-language agents to search and reason over difficult-to-access information. 

Nonetheless, \cite{wei2025browsecomp} primarily operates within the English-language web, missing the linguistic, structural, and cultural complexities inherent to other language environments. In particular, the Chinese web poses unique challenges for retrieval and reasoning: information is scattered across heterogeneous platforms (e.g., Baidu Baike, Zhihu, government portals), naming conventions are inconsistent, and search engines often fail to reliably index deep or domain-specific content. Moreover, the linguistic properties of Chinese, such as implicit referents, idiomatic expressions, and context-dependent syntax, frequently break standard keyword-based retrieval paths. 

Crucially, \textbf{directly translating English browsing benchmarks into Chinese does not yield a meaningful evaluation framework}. This approach fails on several fronts: structural and idiomatic differences render many translated queries unnatural or ineffective in actual search; canonical information pathways in English (e.g., Wikipedia, IMDb) lack equivalents or exhibit vastly different structures in the Chinese web; and translated queries often collapse into keyword matches, trivializing the intended challenge. Moreover, the Chinese web presents its own set of retrieval obstacles—fragmented and platform-specific content, inconsistent naming conventions, and linguistic traits such as ellipsis, cultural references, and implicit reasoning that undermine linear search strategies. As a result, existing English-centric benchmarks struggle to generalize, leaving a critical gap in assessing real-world browsing capabilities in non-English environments. We argue that \textbf{web-based benchmarks must be natively constructed within the Chinese information ecosystem}, where the search logic, content structure, and linguistic context authentically reflect the challenges faced by agents operating in Chinese-language settings.

\begin{figure}[ht]
  \centering
  \includegraphics[width=1\textwidth]{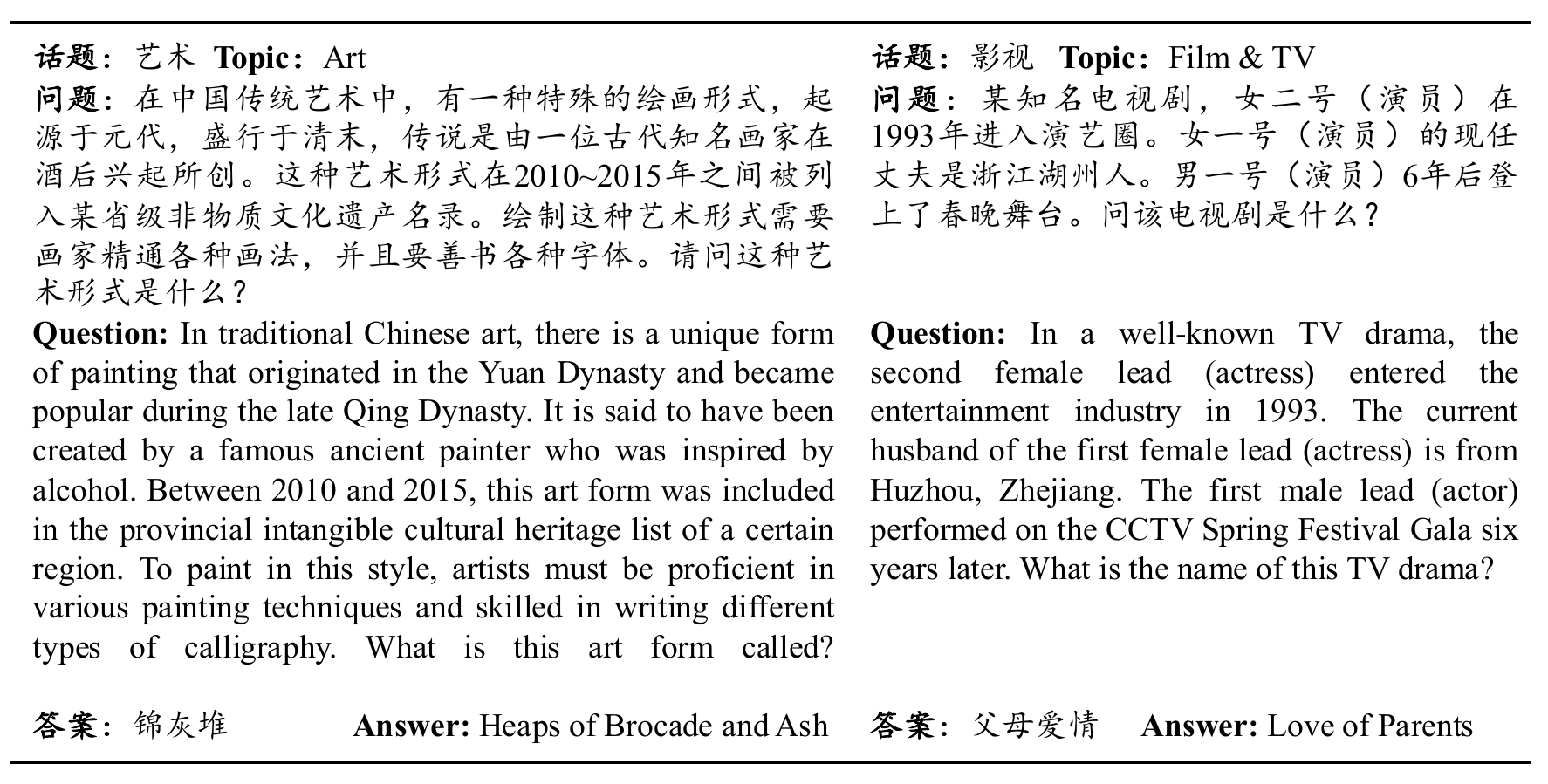}
  \caption{Two data samples from BrowseComp-ZH with their English translations.}
  \label{fig:data_sample}
\end{figure}

To fill this gap, we introduce \textbf{BrowseComp-ZH}, the first benchmark specifically designed to evaluate web-enabled LLMs in the Chinese information environment. Mirroring the philosophy of the original BrowseComp, each item is created by reverse design: expert annotators (all holding at least a master’s degree and possessing both LLM expertise and topical domain knowledge) begin with a single, factual answer and craft a multi-constraint query that is hard to retrieve yet trivially verifiable.
BrowseComp-ZH extends the original framework in two crucial directions:
(1) Native Chinese construction. All queries, evidence chains, and browsing steps are authored in Chinese and fully localized. Every candidate query is verified on three widely-used search engines: \href{https://www.baidu.com}{Baidu}\footnote{\url{https://www.baidu.com}}, \href{https://cn.bing.com}{Bing}\footnote{\url{https://cn.bing.com}}, and \href{https://www.google.com}{Google}\footnote{\url{https://www.google.com}}, and is accepted only if the answer is not surfaced in the first-page results of any engine.
(2) Two-stage quality control. Stage 1 screens for keyword retrievability as above. Stage 2 focuses on ensuring the uniqueness of the answers. We employ a human-in-the-loop approach, where multiple top-performing AI agents first perform reasoning and provide answers based on the designed questions. These results are then manually verified by annotators to check if any additional answers satisfy the constraints of the question.

The final \textbf{BrowseComp-ZH} dataset consists of 289 complex questions, each with multiple constraints and unique answers, spanning 11 diverse domains, including film, technology, law, and medicine. These questions require multi-step reasoning and are difficult to answer directly through search engines, making them ideal for evaluating the ability of Chinese LLM agents to conduct multi-hop retrieval, perform factual reasoning, and integrate online information.
Fig.~\ref{fig:data_sample} illustrates two representative example drawn from BrowseComp-ZH.
Based on this dataset, we benchmark the performance of \textbf{more than 20 systems}, encompassing open-source models (e.g., DeepSeek R1 \cite{guo2025deepseek}, Qwen-2.5-72B-Instruct \cite{yang2024qwen2}), closed-source APIs (e.g., GPT-4o \cite{gpt4o}, Claude-3.7 \cite{claude_3_7}, Gemini-2.5-Pro \cite{gemini_2_5}), as well as AI search agents (e.g., DeepResearch \cite{openai_deep_research}, DeepSeek \cite{deepseek}, Perplexity \cite{perplexity}, and Doubao \cite{doubao}). 

Our evaluation offers a nuanced characterization of model capabilities across different system designs. Based on our analysis, we highlight several key findings:
\begin{enumerate}
    \item Naive large language models, irrespective of parameter scale or training data size, exhibit consistently poor performance on the benchmark, with accuracies often remaining below 10\%. Representative examples include Qwen2.5-72B-Instruct, Llama 4, and GPT-4o.
    
    \item Models endowed with inherent reasoning capabilities exhibit consistent improvements in accuracy. For example, DeepSeek-R1 surpasses DeepSeek-V3 by 14.5\%, while Claude-3.7-Sonnet outperforms Claude-3.5 by 12.2\%.
    
    \item Agentic systems augmented with external retrieval mechanisms tend to achieve comparatively higher scores. Notably, OpenAI's DeepResearch and Doubao (Deep Search) demonstrate that well-orchestrated retrieval and reasoning pipelines can substantially enhance performance, achieving accuracies of 42.9\% and 26.0\%, respectively.

    \item While well-designed retrieval pipelines can significantly improve performance, not all systems benefit from retrieval integration. For instance, enabling web search for DeepSeek-R1 results in a substantial decline in performance, with accuracy dropping from 23.2\% in the direct-answer (no web access) setting to 7.6\% when web search is enabled.

\end{enumerate}

These findings suggest that BrowseComp-ZH not only challenges models' browsing and reasoning capabilities but also exposes their limitations in processing, evaluating, and aligning retrieved information with internal representations. Detailed performance statistics across model categories are summarized in Table~\ref{tab:performance}.

\section{Related Work}
With the increasing ability of large language models (LLMs) to use external tools, recent research has focused on evaluating their capacity to retrieve and reason over real-world information. Representative works such as WebGPT~\cite{nakano2021webgpt}, Toolformer~\cite{schick2023toolformer}, and ReAct~\cite{yao2023react} explore how LLMs leverage search engines, tool usage, and reasoning strategies to tackle complex question-answering tasks. In parallel, retrieval-augmented generation (RAG) frameworks~\cite{guu2020retrieval, lewis2020retrieval} have been widely adopted in QA~\cite{wiratunga2024cbr}, summarization~\cite{edge2024local}, and fact verification~\cite{martin2024semantic}, serving as a mechanism for injecting external knowledge into LLMs.

To assess retrieval capabilities, a variety of widely used English benchmarks have been proposed, including TriviaQA~\cite{joshi2017triviaqa}, HotpotQA~\cite{yang2018hotpotqa}, FEVER~\cite{thorne2018fever}, KILT~\cite{petroni2020kilt}, and GAIA~\cite{mialon2023gaia}. These datasets cover multi-hop reasoning, knowledge-intensive QA, and fact checking, typically relying on structured sources like Wikipedia and StackExchange. However, since many answers can be retrieved via simple keyword searches, these benchmarks often fail to evaluate an agent’s ability to plan complex search trajectories and synthesize information across documents.~\cite{wei2025browsecomp} addresses this by reverse-designing queries from known answers with multiple retrieval constraints, requiring multi-hop search and cross-page reasoning. While it offers finer-grained evaluation for web browsing agents, it remains confined to the English web and lacks generalizability to non-English environments with fragmented platforms and diverse linguistic structures.

However, extending such evaluations to Chinese poses unique challenges. Several retrieval-related datasets have emerged for the Chinese web, but each presents notable limitations.~\cite{hu2024level} evaluates Chinese web search agents but lacks rigorous control over task difficulty and answer accessibility.~\cite{xu2024let} focuses on dynamic, time-sensitive QA tasks but does not emphasize multi-hop retrieval.~\cite{lyu2025crud} evaluates RAG systems under the CRUD (Create, Read, Update, Delete) paradigm, yet primarily emphasizes generation quality over retrieval path validation.~\cite{liu2023benchmarking} and~\cite{li2023huatuo} focus on domain-specific medical QA, but do not evaluate open-ended retrieval or browsing strategies.

To address these limitations in Chinese retrieval benchmarking, BrowseComp-ZH introduces three key innovations: (1) reverse-designed tasks in native Chinese to avoid translation artifacts; (2) rigorous multi-step validation to ensure high retrieval difficulty and answer verifiability; and (3) broad model coverage for evaluating both open-source and proprietary agents. It establishes a high-difficulty benchmark for Chinese web retrieval, supporting systematic evaluation of agents’ ability to navigate fragmented, unstructured, and linguistically diverse Chinese information sources.

\section{The BrowseComp-ZH Dataset}

\subsection{Dataset Construction}

Inspired by \citet{wei2025browsecomp}, we adopt a reverse construction strategy: each task begins with a factual answer, from which an elaborate, multi-constraint query is crafted to make direct retrieval non-trivial. To ensure high-quality annotation, we recruited 10 expert contributors (with Master's or PhD degrees) who have extensive experience in both LLM usage and web search. The overall process comprises following two stages: 

\paragraph{Stage 1: Topic and Answer Selection}
Each annotator selects at least 5 topics from a predefined list spanning \textit{Film \& TV}, \textit{Technology}, \textit{Art}, \textit{History}, \textit{Sports}, \textit{Music}, \textit{Geography}, \textit{Policy \& Law}, \textit{Medicine}, \textit{Video Games}, and \textit{Academic Research}, based on their personal interests. For each topic, they identify several factual answers (e.g., person names, dates, titles, institutions) that meet two criteria: (1) The selected facts must be objective statements that can be independently verified through reliable sources, without the need for interpretation or inference; and (2) they must be concrete and specific enough to exclude overly generic or widely known common-sense facts.

\paragraph{Stage 2: Reverse Question Design}

Building on the selected factual answers, annotators construct complex queries that require integrating contextual cues and external knowledge. The design follows three key principles:

\begin{itemize}[labelindent=\parindent,leftmargin=10pt, itemsep=0pt, parsep=0pt]
\item \textbf{Multi-constraint design}: Each question combines temporal, spatial, categorical, or descriptive conditions to ensure answer uniqueness; not all conditions are required, but at least two dimensions are typically combined;
\item \textbf{Non-trivial retrieval}: Annotators test each query on Baidu, Bing, and Google, using three distinct keyword combinations per search engine. If the correct answer appears on the first page of any search engine, the query will be revised or further constrained;
\item \textbf{Evidence traceability}: Each sample includes at least one authoritative source URL that validates the logical connection between the query constraints and the target answer.
\end{itemize}

To further discourage shortcut-based resolution, all questions are tested using GPT-4o and DeepSeek, both operating in web-enabled search mode. If both models consistently retrieve the correct answer with minimal effort, the query will be revised to increase its complexity by methods including introducing implicit constraints or obfuscating key lexical signals. This process yields 480 preliminary samples, which are subsequently reviewed via the quality control procedure detailed in Section~\ref{sec:quality-control}.

\subsection{Quality Control}
\label{sec:quality-control}

To ensure the rigor and challenge of BrowseComp-ZH, we implement a two-stage quality control protocol: one focusing on question difficulty and the other on answer uniqueness.

\paragraph{Stage 1: Question Difficulty Validation}

Although annotators are required to check whether a question can be quickly solved by search engines during the design process, variations in search ability and prior knowledge could introduce inconsistencies, leading to the inclusion of overly simple questions. To address this, we conduct a cross-checking phase:

\begin{itemize}[labelindent=\parindent,leftmargin=10pt, itemsep=0pt, parsep=0pt]
\item Each annotator validates questions written by others using only search engines (no LLMs);
\item A strict 10-minute time limit is applied per question;
\item If the answer is found within the time limit, the task is labeled \emph{low difficulty};
\item If the answer is not found and the question structure is logical and verifiable, it is labeled \emph{high difficulty}.
\end{itemize}

In this stage, annotators identified 76 simple samples, and after filtering them out, we are left with 404 high-difficulty candidates.

\paragraph{Stage 2: Answer Uniqueness Validation}

This stage focuses on ensuring the uniqueness of the answers, meaning that there is only one correct and unambiguous answer that satisfies all the constraints of each question. We employ a human-in-the-loop approach as follows:

\begin{itemize}[labelindent=\parindent,leftmargin=10pt, itemsep=0pt, parsep=0pt]
\item Multiple top-performing AI agents refer to the models that perform best on the original 404 high-difficulty candidates based on their ability to reason and retrieve accurate answers.
 AI agents, including OpenAI DeepSearch, Perplexity, Doubao (with deep reasoning), OpenAI O1, and Gemini 2.5-Pro, first generate reasoning processes and answers for each question.
\item These results are then manually verified by annotators to check if any alternative answers satisfy the constraints of the question.
\item If any alternative answer meets all task constraints (e.g., factual accuracy, specificity, verifiability) but differs from the original answer, the task is considered ambiguous and rejected.
\end{itemize}

This process eliminates 115 ambiguous samples, resulting in a final benchmark of 289 validated questions that maintains high levels of difficulty and answer verifiability.

\subsection{Data Statistics}

\begin{figure}[t]
  \centering
  \includegraphics[width=1\textwidth]{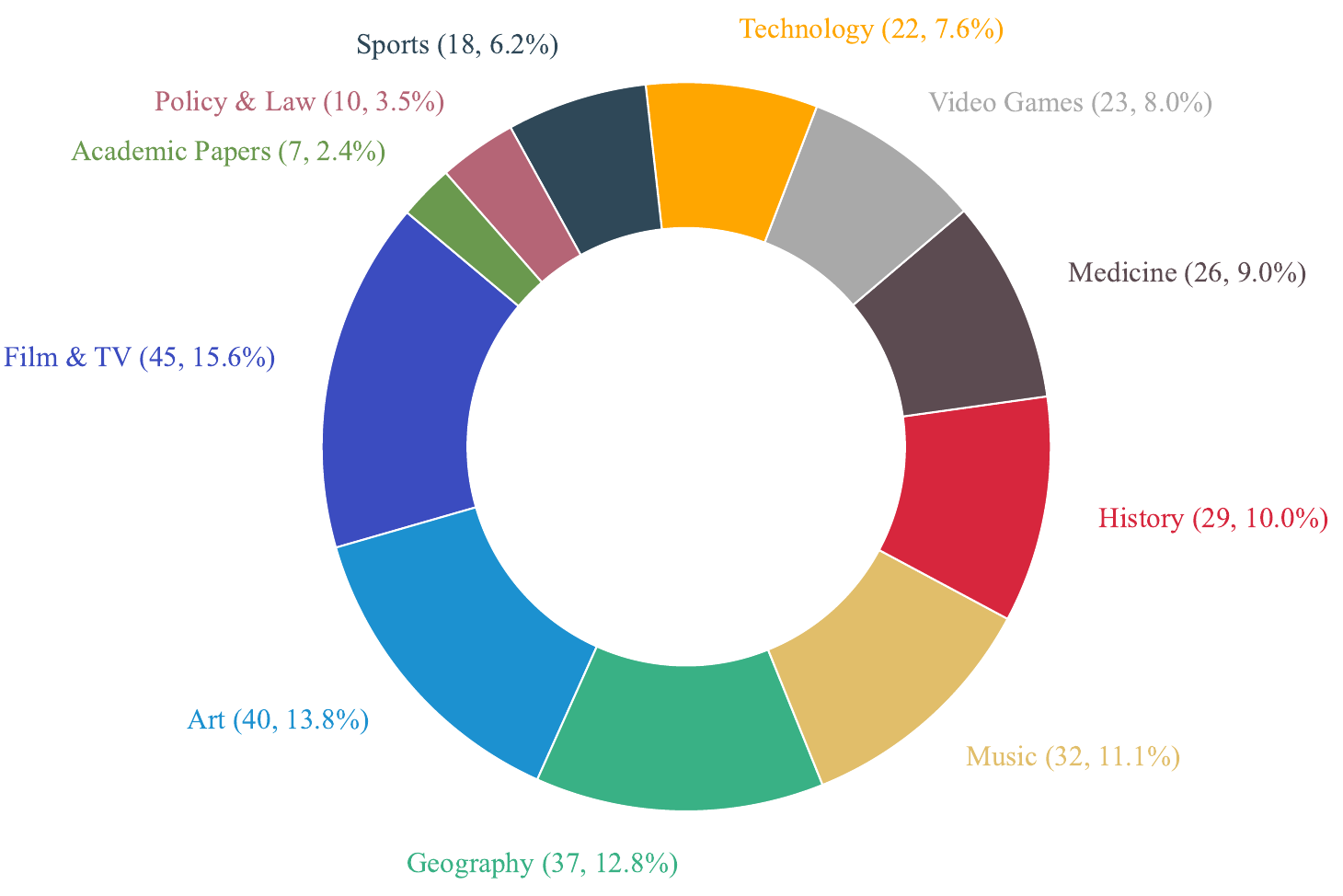}
  \caption{Distribution of samples across 11 topic domains in BrowseComp-ZH dataset.}
  \label{fig:topic_distribution}
\end{figure}

\begin{figure}[ht]
  \centering
  \includegraphics[width=1\textwidth]{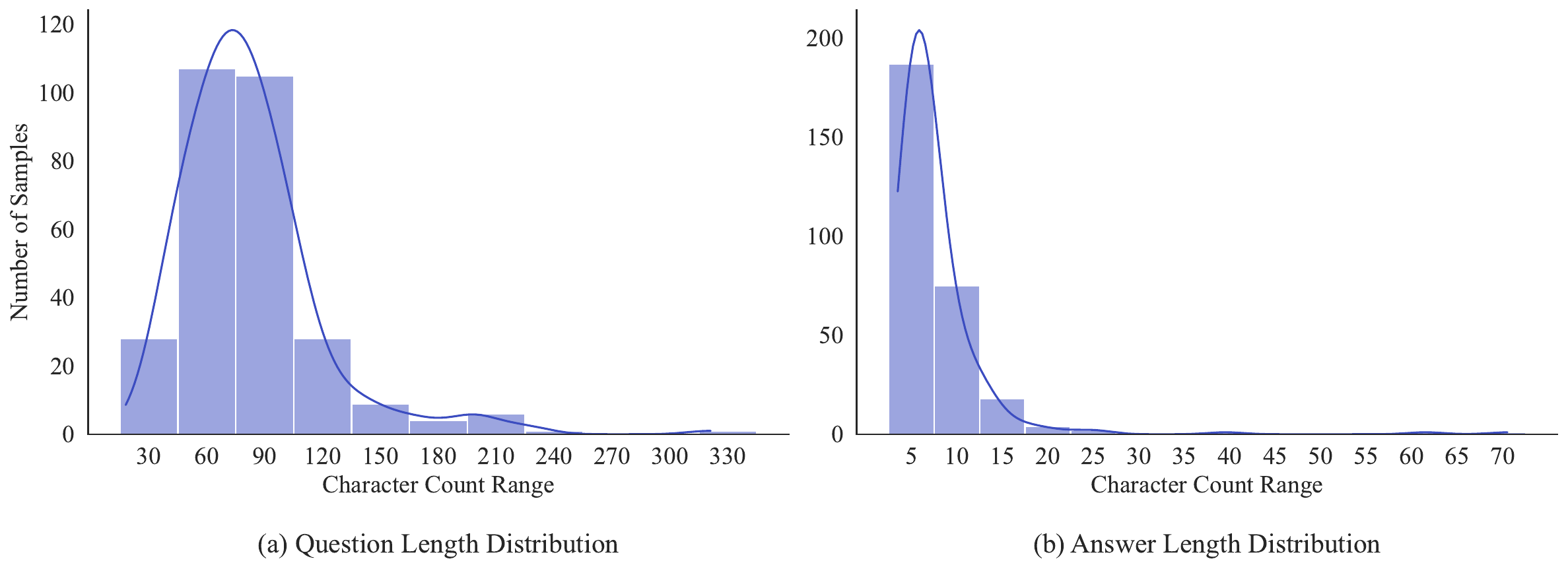}
  \caption{Distribution of question and answer lengths in the BrowseComp-ZH dataset.}
  \label{fig:length_distribution}
\end{figure}

In this section, we present statistics on the topic distribution, question and answer length distribution of our curated BrowseComp-ZH dataset.

\textbf{Topic distribution.} Fig.~\ref{fig:topic_distribution}  presents the distribution of samples across 11 topic domains in the BrowseComp-ZH dataset. As illustrated, the most represented categories include \textit{Film \& TV} (15.6\%), \textit{Art} (13.8\%), and \textit{Geography} (12.8\%), reflecting the diverse interests of annotators and a broad coverage of Chinese web content. The dataset also features \textit{Music} (11.1\%), \textit{History}(10.0\%), and \textit{Medicine} (9.0\%). Conversely, topics like \textit{Policy \& Law} (3.5\%) and \textit{Academic Papers} (2.4\%) have fewer samples, likely due to the complexity of sourcing factual answers from these areas. The distribution underscores the multi-disciplinary nature of the dataset, aimed at evaluating language models across a wide array of knowledge domains.

\textbf{QA Length distribution.} Fig.~\ref{fig:length_distribution} illustrates the distribution of question and answer lengths in the BrowseComp-ZH dataset. As shown in Fig.~\ref{fig:length_distribution} (a), the question length predominantly ranges between 60 to 90 characters, with the majority of questions falling within this interval. This suggests that questions are designed to be succinct but information-dense, requiring substantial detail to challenge the models. In Fig.~\ref{fig:length_distribution} (b), the answer length distribution shows that answers are generally short, typically falling within the range of 5 to 10 characters. This aligns with the design principle of providing precise and verifiable answers, ensuring that the responses are direct and concise. These length distributions reflect the high information density of the dataset, emphasizing both complexity in the questions and simplicity in the answers.

\section{Benchmarks}
\subsection{Models}
We evaluate a wide range of state-of-the-art open-source and proprietary models, as well as mainstream AI search products on BrowseComp-ZH, aiming to provide a diverse, comprehensive, and instructive benchmark.

\begin{itemize}
\item \textbf{Open-source models:} DeepSeek-V3 \cite{liu2024deepseek}, DeepSeek-R1 \cite{guo2025deepseek}, Qwen2.5-72B-Instruct \cite{yang2024qwen2}, Qwen3-235B-A22B \cite{yang2024qwen2}, QwQ-32B \cite{qwq-32b-preview}, and LlaMa4 (maverick-instruct-basic) \cite{llama4}.
\item \textbf{Closed-source models:} GPT-4o \cite{gpt4o}, O1 \cite{o1}, O4-mini \cite{o4-mini}, Claude-3.5-Sonnet() \cite{claude_3_5}, Claude-3.7-Sonnet(20250219) \cite{claude_3_7}, Gemini-2.0-Flash() \cite{gemini_2_0}, Gemini-2.5-Pro(preview-03-25) \cite{gemini_2_5}, and Qwen2.5-MAX (qwen-max-2025-01-25) \cite{qwen_2_5_max}
\item \textbf{AI search products:} DeepResearch \cite{openai_deep_research}, Grok-3 \cite{grok3}, Perplexity (Research mode) \cite{perplexity}, Doubao (with both deep search and standard modes) \cite{doubao}, Kimi (deep think version) \cite{kimi}, Yuanbao (Hunyuan Model) \cite{yuanbao}, and DeepSeek (with both deep think and standard modes) \cite{deepseek}.
\end{itemize}

\subsection{Setting}
For models such as O1 and Claude-3.7, inference is performed with fixed temperature and top-p settings, as these parameters are not user-configurable. For models that support custom decoding parameters, we follow the configuration used in DeepSeek R1 \cite{guo2025deepseek}, setting temperature to 0.6 and top-p to 0.95. For AI search products, we recruited human annotators to perform GUI-based interactions and recorded the outputs for subsequent analysis.

Since the questions in BrowseComp-ZH benchmark are designed to elicit concise factual answers, evaluating the correctness of model outputs is straightforward. For both open-source and closed-source models, which are generally capable of accurately following the provided instructions, we extract answers from the model responses using regular expressions and score them using GPT-4o. The grading prompts are adapted from those used in the original BrowseComp benchmark \cite{wei2025browsecomp}. Full grading prompts are included in Appendix \ref{fig:judge_prompt}.

In contrast, AI search products typically undergo additional post-training to align with product-specific features, which often reduces their ability to follow instructions precisely. Therefore, we employ human annotators to manually extract the answers from these systems and verify their consistency with the ground truth.

The instructions provided to both open-source and closed-source models, as well as AI search products, are detailed in Appendix~\ref{fig:instruction}.

\subsection{Metrics}
To quantitatively evaluate model performance, we report both accuracy and calibration error. For the calculation of calibration error, we partition the predicted probabilities into five bins: [0–0.2), [0.2–0.4), [0.4–0.6), [0.6–0.8), and [0.8–1.0]. For each bin, we compute the absolute difference between the average predicted probability and the actual accuracy, and then calculate a weighted average across all bins. The ECE is formally defined as:
$$
\mathrm{ECE}=\sum_{i=1}^B \frac{n_i}{N}|acc(i)-conf(i)|
$$
where $B$ is the number of bins, $n_i$ is the number of samples in the $i$-th bin. $acc(i)$ is the empirical accuracy of the $i$-th bin, $conf(i)$ is the average predicted confidence in the $i$-th bin.

\subsection{Performance}
As shown in the performance comparison of models and AI search products on our benchmark (Tab.~\ref{tab:performance}), several systems, such as Qwen2.5-72B-Instruct (6.6\% accuracy), GPT-4o (6.2\%), and Claude-3.5-Sonnet (5.5\%), exhibited limited performance, consistently achieving low accuracy scores across tasks, highlighting the challenging nature of the dataset we proposed. Among all evaluated systems, OpenAI's DeepResearch achieved the highest accuracy (42.9\%), followed by OpenAI's O1 (29.1\%) and Google's Gemini-2.5-Pro (27.3\%). Interestingly, even models without browsing capabilities, such as DeepSeek R1, O1, and Gemini-2.5-Pro, which rely solely on their internal world knowledge, managed to achieve accuracies exceeding 20\%, demonstrating the strength of high-capacity language models. Overall, AI search products equipped with retrieval mechanisms outperformed other categories, followed by closed-source APIs, while open-source models performed the worst on this challenging benchmark. These results demonstrate the effectiveness of our benchmark in differentiating models across a wide range of performance levels.

\begin{table}[]
\resizebox{\textwidth}{!}{
\begin{tabular}{@{}lcccccc@{}}
\toprule
Model                   & Category          & Reasoning & Browsing & Accuracy & Calibration Error(\%) & Enterprise \\ \midrule
DeepSeek-V3             & Open-Source       & N         & N        & 8.7\%    & 72                    & DeepSeek   \\
DeepSeek-R1             & Open-Source       & Y         & N        & 23.2\%   & 59                    & DeepSeek   \\
Qwen2.5-72B-Instruct    & Open-Source       & N         & N        & 6.6\%    & 62                    & Alibaba    \\
Qwen3-235B-A22B(Non-Thinking)    & Open-Source       & N         & N        & 8.0\%    & 80                    & Alibaba    \\
Qwen3-235B-A22B(Thinking)    & Open-Source       & Y         & N        & 13.2\%    & 67                    & Alibaba    \\
QwQ-32B                 & Open-Source       & Y         & N        & 11.1\%   & 64                    & Alibaba    \\
LlaMa4                  & Open-Source       & N         & N        & 4.8\%    & 70                    & Meta       \\ \midrule
GPT4o                   & Closed-Source     & N         & N        & 6.2\%    & 73                    & OpenAI     \\
O1                      & Closed-Source     & Y         & N        & \underline{29.1\%}   & 52                    & OpenAI     \\
O4-mini                 & Closed-Source     & Y         & N        & 15.2\%   & 42                    & OpenAI     \\
Claude-3.5-Sonnet       & Closed-Source     & N         & N        & 5.5\%    & 78                    & Anthropic  \\
Claude-3.7-Sonnet       & Closed-Source     & Y         & N        & 17.7\%   & 71                    & Anthropic  \\
Gemini-2.0-Flash        & Closed-Source     & N         & N        & 6.9\%    & 74                    & Google     \\
Gemini-2.5-Pro          & Closed-Source     & Y         & N        & 27.3\%   & 59                    & Google     \\
Qwen2.5-MAX             & Closed-Source     & N         & N        & 7.6\%    & 78                    & Alibaba    \\\midrule
OpenAI DeepResearch     & AI Search Product & -         & Y        & \textbf{42.9\%}   & 9                     & OpenAI     \\
Grok3 (Research)        & AI Search Product & -         & Y        & 12.9\%   & 39                    & xAI        \\
Perplexity (Research)   & AI Search Product & -         & Y        & 22.6\%   & 53                    & Perplexity \\
Doubao (Deep Search)    & AI Search Product & -         & Y        & 26.0\%   & 61                    & ByteDance  \\
Doubao (Standard)       & AI Search Product & -         & Y        & 18.7\%   & 37                    & ByteDance  \\
Kimi (Deep Think)       & AI Search Product & -         & Y        & 8.0\%    & 58                    & Moonshot       \\
Yuanbao (Hunyuan Model) & AI Search Product & -         & Y        & 12.2\%   & 56                    & Tencent    \\
DeepSeek (Deep Think)   & AI Search Product & -         & Y        & 7.6\%    & 65                    & DeepSeek   \\
DeepSeek (Standard)     & AI Search Product & -         & Y        & 4.8\%    & 66                    & DeepSeek   \\ \bottomrule
\end{tabular}
}
\vspace{3mm}
\caption{Performance of various models and AI search products on BrowseComp-ZH dataset.}
\label{tab:performance}
\end{table}

\subsection{Analysis of Reasoning Ability of LLM}
Since the tasks in our proposed BrowseComp-ZH benchmark require both extensive world knowledge and strong reasoning abilities, we further investigate the impact of reasoning on model performance. As summarized in Tab.~\ref{tab:performance}, we compare models with and without explicit reasoning mechanisms, including DeepSeek-V3 versus DeepSeek-R1, Claude-3.5-Sonnet versus Claude-3.7-Sonnet, and Gemini-2.0-Flash versus Gemini-2.5-Pro.
Across all comparisons, models enhanced with reasoning capabilities consistently demonstrate substantial performance gains. For example, DeepSeek-R1 achieves an accuracy of 23.2\%, markedly improving upon DeepSeek-V3’s 8.7\%, while Claude-3.7-Sonnet outperforms Claude-3.5-Sonnet (17.7\% vs. 5.5\%).

A similar pattern emerges among AI search products. Doubao (Deep Search) achieves a nearly 8\% absolute improvement over Doubao (Standard) (26.0\% vs. 18.7\%), highlighting the benefits of enhanced reasoning in facilitating more accurate and iterative retrieval for complex queries.

However, this trend is less evident in DeepSeek's AI search products. Notably, all versions of DeepSeek’s search system perform only a single round of retrieval, limiting the extent to which improved reasoning capabilities can be leveraged. Consequently, DeepSeek’s Deep Search variant does not exhibit a significant performance advantage over its standard counterpart.

\subsection{Analysis of AI Search Products}
Currently, AI search systems can be broadly categorized into two types: those that perform a single retrieval to answer a user's query (e.g., Kimi, Tencent Yuanbao based on the Hunyuan model, and DeepSeek), and those that conduct multiple rounds of retrieval, iteratively refining or expanding the search based on the query and intermediate results (e.g., DeepResearch, Perplexity, and Doubao).
Statistical analysis shows that systems employing multi-round retrieval achieve significantly higher accuracy, with DeepResearch reaching 42.9\%, Doubao (Deep Search) achieving 26.0\%, and Perplexity attaining 22.6\%, compared to single-retrieval systems such as Kimi (8.0\%), Yuanbao (12.2\%), and DeepSeek (7.6\%).
This trend aligns with the nature of tasks in BrowseComp-ZH, which often involve multi-faceted queries, making it challenging to obtain accurate answers through a single retrieval operation.

Notably, we observe a counterintuitive phenomenon: enabling search functionality for DeepSeek-R1 leads to a substantial decline in performance, with accuracy dropping from 23.2\% in the direct-answer (no web access) setting to 7.6\% when web search is enabled.
We hypothesize that this degradation arises because, without effective alignment mechanisms, the model may rely on less reliable retrieved content, which in turn overrides its more accurate internal knowledge.

This observation highlights a critical challenge for large language models: effectively reconciling retrieved evidence with internal representations remains non-trivial. Furthermore, integrating retrieval capabilities without robust post-retrieval reasoning and alignment strategies may, in some cases, hinder rather than enhance model performance.

\subsection{Calibration Analysis}
We also evaluate the model’s calibration, which measures the alignment between predicted confidence scores and actual accuracy. Following the methodology adopted in BrowseComp, we require models to provide confidence estimates alongside their predictions during evaluation.
As shown in Tab.~\ref{tab:performance}, integrating search functionality results in increased calibration errors. For instance, the calibration error for DeepSeek-R1 increases from 59\% in the direct-answer setting to 65\% when search is enabled. This trend is consistent with the observations reported in BrowseComp.

\section{Conclusion and Discussion}

This study introduces BrowseComp-ZH, the first benchmark specifically designed to evaluate the web browsing and reasoning capabilities of large language models (LLMs) in the Chinese information environment. Inspired by the \textsc{BrowseComp} benchmark, we construct challenging question-answer pairs that require multi-hop retrieval, information filtering, and logical reasoning to derive concise, factual answers. To ensure the high difficulty of each question and the uniqueness of each answer, we implement a rigorous two-stage quality control pipeline that includes a three-engine keyword validation process and human-in-the-loop verification, ensuring that answers are both difficult to retrieve and unambiguous.

We construct 289 high-quality samples across 11 diverse topics, including \textit{Film \& TV}, \textit{History}, \textit{Technology}, \textit{Medicine}, and more. Using these tasks, we conduct extensive evaluations on over 20 models and AI search products, encompassing open-source LLMs, closed-source APIs, and AI search products. As shown in Tab.~\ref{tab:performance}, most standalone LLMs—such as GPT-4o, Qwen2.5-72B, and Llama-4—achieve limited accuracy, highlighting the difficulty of the benchmark. Models with stronger reasoning abilities, such as O1 (29.1\%) and Gemini-2.5-Pro (27.3\%), demonstrate substantial improvements, underscoring the critical role of reasoning for complex question answering. AI search products employing multi-turn retrieval, including DeepResearch (42.9\%) and Doubao (Deep Search) (26.0\%), further outperform purely parametric models, illustrating the effectiveness of test-time scaling through iterative retrieval. These results reflect the inherent challenges of the Chinese web environment, where fragmented information and inconsistent indexing complicate single-shot search strategies.

\textbf{Limitations.} Despite the innovations in BrowseComp-ZH's design and quality control, several limitations remain. First, the current dataset is relatively small; increasing both the sample size and the diversity of question types would improve its representativeness.  Second, although we apply rigorous validation to ensure answer uniqueness, it cannot be fully guaranteed. In particular, the dynamic nature of the web means that factual answers may evolve or become inconsistent over time, making stability and reproducibility an ongoing challenge.

% \paragraph{Ethics \& Societal Impacts}

% The data used in this study is sourced from publicly accessible Chinese web content, ensuring the ethical use of publicly available information without infringing on individual privacy. We emphasize that \textbf{BrowseComp-ZH} is intended strictly for academic research purposes and should not be used for decision-making tasks in sensitive domains such as healthcare or legal affairs. We also highlight the importance of recognizing the limitations of the dataset when employing models trained on it, especially in contexts that require ethical considerations.

\paragraph{Future Work}
In future work, we plan to incorporate a broader range of questions to enable more comprehensive and accurate evaluations of the models. We will also conduct an in-depth analysis of their reasoning mechanisms and search strategies. Furthermore, additional case studies will be carried out to examine failure cases across different models. Finally, we aim to explore methods to further improve the models’ browsing and reasoning capabilities, such as leveraging post-training techniques like reinforcement learning.

\bibliographystyle{ACM-Reference-Format}
\bibliography{nips}


\begin{thebibliography}{48}

%%% the Web address.

\ifx \showCODEN    \undefined \def \showCODEN     #1{\unskip}     \fi
\ifx \showDOI      \undefined \def \showDOI       #1{#1}\fi
\ifx \showISBNx    \undefined \def \showISBNx     #1{\unskip}     \fi
\ifx \showISBNxiii \undefined \def \showISBNxiii  #1{\unskip}     \fi
\ifx \showISSN     \undefined \def \showISSN      #1{\unskip}     \fi
\ifx \showLCCN     \undefined \def \showLCCN      #1{\unskip}     \fi
\ifx \shownote     \undefined \def \shownote      #1{#1}          \fi
\ifx \showarticletitle \undefined \def \showarticletitle #1{#1}   \fi
\ifx \showURL      \undefined \def \showURL       {\relax}        \fi
% invisible to TeX
\providecommand\bibfield[2]{#2}
\providecommand\bibinfo[2]{#2}
\providecommand\natexlab[1]{#1}
\providecommand\showeprint[2][]{arXiv:#2}

\bibitem[Anthropic(2024)]%
        {claude_3_5}
\bibfield{author}{\bibinfo{person}{Anthropic}.} \bibinfo{year}{2024}\natexlab{}.
\newblock \bibinfo{title}{Introducing Claude 3.5 Sonnet}.
\newblock \bibinfo{howpublished}{\url{https://www.anthropic.com/news/claude-3-5-sonnet}}.
\newblock


\bibitem[Anthropic(2025)]%
        {claude_3_7}
\bibfield{author}{\bibinfo{person}{Anthropic}.} \bibinfo{year}{2025}\natexlab{}.
\newblock \bibinfo{title}{Introducing Claude 3.7 Sonnet and Claude Code}.
\newblock \bibinfo{howpublished}{\url{https://www.anthropic.com/news/claude-3-7-sonnet}}.
\newblock


\bibitem[ByteDance(2025)]%
        {doubao}
\bibfield{author}{\bibinfo{person}{ByteDance}.} \bibinfo{year}{2025}\natexlab{}.
\newblock \bibinfo{howpublished}{\url{https://www.doubao.com/chat/}}.
\newblock


\bibitem[DeepSeek(2025)]%
        {deepseek}
\bibfield{author}{\bibinfo{person}{DeepSeek}.} \bibinfo{year}{2025}\natexlab{}.
\newblock \bibinfo{howpublished}{\url{https://chat.deepseek.com/}}.
\newblock


\bibitem[Edge et~al\mbox{.}(2024)]%
        {edge2024local}
\bibfield{author}{\bibinfo{person}{Darren Edge}, \bibinfo{person}{Ha Trinh}, \bibinfo{person}{Newman Cheng}, \bibinfo{person}{Joshua Bradley}, \bibinfo{person}{Alex Chao}, \bibinfo{person}{Apurva Mody}, \bibinfo{person}{Steven Truitt}, \bibinfo{person}{Dasha Metropolitansky}, \bibinfo{person}{Robert~Osazuwa Ness}, {and} \bibinfo{person}{Jonathan Larson}.} \bibinfo{year}{2024}\natexlab{}.
\newblock \showarticletitle{From local to global: A graph rag approach to query-focused summarization}.
\newblock \bibinfo{journal}{\emph{arXiv preprint arXiv:2404.16130}} (\bibinfo{year}{2024}).
\newblock


\bibitem[Fan et~al\mbox{.}(2024)]%
        {fan2024survey}
\bibfield{author}{\bibinfo{person}{Wenqi Fan}, \bibinfo{person}{Yujuan Ding}, \bibinfo{person}{Liangbo Ning}, \bibinfo{person}{Shijie Wang}, \bibinfo{person}{Hengyun Li}, \bibinfo{person}{Dawei Yin}, \bibinfo{person}{Tat-Seng Chua}, {and} \bibinfo{person}{Qing Li}.} \bibinfo{year}{2024}\natexlab{}.
\newblock \showarticletitle{A survey on rag meeting llms: Towards retrieval-augmented large language models}. In \bibinfo{booktitle}{\emph{Proceedings of the 30th ACM SIGKDD Conference on Knowledge Discovery and Data Mining}}. \bibinfo{pages}{6491--6501}.
\newblock


\bibitem[Fern{\'a}ndez-Pichel et~al\mbox{.}(2024)]%
        {fernandez2024search}
\bibfield{author}{\bibinfo{person}{Marcos Fern{\'a}ndez-Pichel}, \bibinfo{person}{Juan~C Pichel}, {and} \bibinfo{person}{David~E Losada}.} \bibinfo{year}{2024}\natexlab{}.
\newblock \showarticletitle{Search Engines, LLMs or Both? Evaluating Information Seeking Strategies for Answering Health Questions}.
\newblock \bibinfo{journal}{\emph{arXiv preprint arXiv:2407.12468}} (\bibinfo{year}{2024}).
\newblock


\bibitem[Google(2024a)]%
        {gemini_2_5}
\bibfield{author}{\bibinfo{person}{Google}.} \bibinfo{year}{2024}\natexlab{a}.
\newblock \bibinfo{title}{Gemini 2.5: Our most intelligent AI model}.
\newblock \bibinfo{howpublished}{\url{https://blog.google/technology/google-deepmind/gemini-model-thinking-updates-march-2025/}}.
\newblock


\bibitem[Google(2024b)]%
        {gemini_2_0}
\bibfield{author}{\bibinfo{person}{Google}.} \bibinfo{year}{2024}\natexlab{b}.
\newblock \bibinfo{title}{Introducing Gemini 2.0: our new AI model for the agentic era}.
\newblock \bibinfo{howpublished}{\url{https://blog.google/technology/google-deepmind/google-gemini-ai-update-december-2024/}}.
\newblock


\bibitem[Guo et~al\mbox{.}(2025)]%
        {guo2025deepseek}
\bibfield{author}{\bibinfo{person}{Daya Guo}, \bibinfo{person}{Dejian Yang}, \bibinfo{person}{Haowei Zhang}, \bibinfo{person}{Junxiao Song}, \bibinfo{person}{Ruoyu Zhang}, \bibinfo{person}{Runxin Xu}, \bibinfo{person}{Qihao Zhu}, \bibinfo{person}{Shirong Ma}, \bibinfo{person}{Peiyi Wang}, \bibinfo{person}{Xiao Bi}, {et~al\mbox{.}}} \bibinfo{year}{2025}\natexlab{}.
\newblock \showarticletitle{Deepseek-r1: Incentivizing reasoning capability in llms via reinforcement learning}.
\newblock \bibinfo{journal}{\emph{arXiv preprint arXiv:2501.12948}} (\bibinfo{year}{2025}).
\newblock


\bibitem[Guu et~al\mbox{.}(2020)]%
        {guu2020retrieval}
\bibfield{author}{\bibinfo{person}{Kelvin Guu}, \bibinfo{person}{Kenton Lee}, \bibinfo{person}{Zora Tung}, \bibinfo{person}{Panupong Pasupat}, {and} \bibinfo{person}{Mingwei Chang}.} \bibinfo{year}{2020}\natexlab{}.
\newblock \showarticletitle{Retrieval augmented language model pre-training}. In \bibinfo{booktitle}{\emph{International conference on machine learning}}. PMLR, \bibinfo{pages}{3929--3938}.
\newblock


\bibitem[He et~al\mbox{.}(2024)]%
        {he2024zero}
\bibfield{author}{\bibinfo{person}{Guangxin He}, \bibinfo{person}{Zonghong Dai}, \bibinfo{person}{Jiangcheng Zhu}, \bibinfo{person}{Binqiang Zhao}, \bibinfo{person}{Qicheng Hu}, \bibinfo{person}{Chenyue Li}, \bibinfo{person}{You Peng}, \bibinfo{person}{Chen Wang}, {and} \bibinfo{person}{Binhang Yuan}.} \bibinfo{year}{2024}\natexlab{}.
\newblock \showarticletitle{Zero-Indexing Internet Search Augmented Generation for Large Language Models}.
\newblock \bibinfo{journal}{\emph{arXiv preprint arXiv:2411.19478}} (\bibinfo{year}{2024}).
\newblock


\bibitem[Hu et~al\mbox{.}(2024)]%
        {hu2024level}
\bibfield{author}{\bibinfo{person}{Chuanrui Hu}, \bibinfo{person}{Shichong Xie}, \bibinfo{person}{Baoxin Wang}, \bibinfo{person}{Bin Chen}, \bibinfo{person}{Xiaofeng Cong}, {and} \bibinfo{person}{Jun Zhang}.} \bibinfo{year}{2024}\natexlab{}.
\newblock \showarticletitle{Level-Navi Agent: A Framework and benchmark for Chinese Web Search Agents}.
\newblock \bibinfo{journal}{\emph{arXiv preprint arXiv:2502.15690}} (\bibinfo{year}{2024}).
\newblock


\bibitem[Joshi et~al\mbox{.}(2017)]%
        {joshi2017triviaqa}
\bibfield{author}{\bibinfo{person}{Mandar Joshi}, \bibinfo{person}{Eunsol Choi}, \bibinfo{person}{Daniel~S Weld}, {and} \bibinfo{person}{Luke Zettlemoyer}.} \bibinfo{year}{2017}\natexlab{}.
\newblock \showarticletitle{Triviaqa: A large scale distantly supervised challenge dataset for reading comprehension}.
\newblock \bibinfo{journal}{\emph{arXiv preprint arXiv:1705.03551}} (\bibinfo{year}{2017}).
\newblock


\bibitem[Lai et~al\mbox{.}(2024)]%
        {lai2024autowebglm}
\bibfield{author}{\bibinfo{person}{Hanyu Lai}, \bibinfo{person}{Xiao Liu}, \bibinfo{person}{Iat~Long Iong}, \bibinfo{person}{Shuntian Yao}, \bibinfo{person}{Yuxuan Chen}, \bibinfo{person}{Pengbo Shen}, \bibinfo{person}{Hao Yu}, \bibinfo{person}{Hanchen Zhang}, \bibinfo{person}{Xiaohan Zhang}, \bibinfo{person}{Yuxiao Dong}, {et~al\mbox{.}}} \bibinfo{year}{2024}\natexlab{}.
\newblock \showarticletitle{AutoWebGLM: A Large Language Model-based Web Navigating Agent}. In \bibinfo{booktitle}{\emph{Proceedings of the 30th ACM SIGKDD Conference on Knowledge Discovery and Data Mining}}. \bibinfo{pages}{5295--5306}.
\newblock


\bibitem[Lai et~al\mbox{.}(2025)]%
        {lai2025webglm}
\bibfield{author}{\bibinfo{person}{Hanyu Lai}, \bibinfo{person}{Xiao Liu}, \bibinfo{person}{Hao Yu}, \bibinfo{person}{Yifan Xu}, \bibinfo{person}{Iat~Long Iong}, \bibinfo{person}{Shuntian Yao}, \bibinfo{person}{Aohan Zeng}, \bibinfo{person}{Zhengxiao Du}, \bibinfo{person}{Yuxiao Dong}, {and} \bibinfo{person}{Jie Tang}.} \bibinfo{year}{2025}\natexlab{}.
\newblock \showarticletitle{WebGLM: Towards an Efficient and Reliable Web-Enhanced Question Answering System}.
\newblock \bibinfo{journal}{\emph{ACM Transactions on Information Systems}} (\bibinfo{year}{2025}).
\newblock


\bibitem[Lewis et~al\mbox{.}(2020)]%
        {lewis2020retrieval}
\bibfield{author}{\bibinfo{person}{Patrick Lewis}, \bibinfo{person}{Ethan Perez}, \bibinfo{person}{Aleksandra Piktus}, \bibinfo{person}{Fabio Petroni}, \bibinfo{person}{Vladimir Karpukhin}, \bibinfo{person}{Naman Goyal}, \bibinfo{person}{Heinrich K{\"u}ttler}, \bibinfo{person}{Mike Lewis}, \bibinfo{person}{Wen-tau Yih}, \bibinfo{person}{Tim Rockt{\"a}schel}, {et~al\mbox{.}}} \bibinfo{year}{2020}\natexlab{}.
\newblock \showarticletitle{Retrieval-augmented generation for knowledge-intensive nlp tasks}.
\newblock \bibinfo{journal}{\emph{Advances in neural information processing systems}}  \bibinfo{volume}{33} (\bibinfo{year}{2020}), \bibinfo{pages}{9459--9474}.
\newblock


\bibitem[Li et~al\mbox{.}(2023a)]%
        {li2023web}
\bibfield{author}{\bibinfo{person}{Junyi Li}, \bibinfo{person}{Tianyi Tang}, \bibinfo{person}{Wayne~Xin Zhao}, \bibinfo{person}{Jingyuan Wang}, \bibinfo{person}{Jian-Yun Nie}, {and} \bibinfo{person}{Ji-Rong Wen}.} \bibinfo{year}{2023}\natexlab{a}.
\newblock \showarticletitle{The web can be your oyster for improving large language models}.
\newblock \bibinfo{journal}{\emph{arXiv preprint arXiv:2305.10998}} (\bibinfo{year}{2023}).
\newblock


\bibitem[Li et~al\mbox{.}(2023b)]%
        {li2023huatuo}
\bibfield{author}{\bibinfo{person}{Jianquan Li}, \bibinfo{person}{Xidong Wang}, \bibinfo{person}{Xiangbo Wu}, \bibinfo{person}{Zhiyi Zhang}, \bibinfo{person}{Xiaolong Xu}, \bibinfo{person}{Jie Fu}, \bibinfo{person}{Prayag Tiwari}, \bibinfo{person}{Xiang Wan}, {and} \bibinfo{person}{Benyou Wang}.} \bibinfo{year}{2023}\natexlab{b}.
\newblock \showarticletitle{Huatuo-26m, a large-scale chinese medical qa dataset}.
\newblock \bibinfo{journal}{\emph{arXiv preprint arXiv:2305.01526}} (\bibinfo{year}{2023}).
\newblock


\bibitem[Liu et~al\mbox{.}(2024)]%
        {liu2024deepseek}
\bibfield{author}{\bibinfo{person}{Aixin Liu}, \bibinfo{person}{Bei Feng}, \bibinfo{person}{Bing Xue}, \bibinfo{person}{Bingxuan Wang}, \bibinfo{person}{Bochao Wu}, \bibinfo{person}{Chengda Lu}, \bibinfo{person}{Chenggang Zhao}, \bibinfo{person}{Chengqi Deng}, \bibinfo{person}{Chenyu Zhang}, \bibinfo{person}{Chong Ruan}, {et~al\mbox{.}}} \bibinfo{year}{2024}\natexlab{}.
\newblock \showarticletitle{Deepseek-v3 technical report}.
\newblock \bibinfo{journal}{\emph{arXiv preprint arXiv:2412.19437}} (\bibinfo{year}{2024}).
\newblock


\bibitem[Liu et~al\mbox{.}(2023)]%
        {liu2023benchmarking}
\bibfield{author}{\bibinfo{person}{Junling Liu}, \bibinfo{person}{Peilin Zhou}, \bibinfo{person}{Yining Hua}, \bibinfo{person}{Dading Chong}, \bibinfo{person}{Zhongyu Tian}, \bibinfo{person}{Andrew Liu}, \bibinfo{person}{Helin Wang}, \bibinfo{person}{Chenyu You}, \bibinfo{person}{Zhenhua Guo}, \bibinfo{person}{Lei Zhu}, {et~al\mbox{.}}} \bibinfo{year}{2023}\natexlab{}.
\newblock \showarticletitle{Benchmarking large language models on cmexam-a comprehensive chinese medical exam dataset}.
\newblock \bibinfo{journal}{\emph{Advances in Neural Information Processing Systems}}  \bibinfo{volume}{36} (\bibinfo{year}{2023}), \bibinfo{pages}{52430--52452}.
\newblock


\bibitem[Lyu et~al\mbox{.}(2025)]%
        {lyu2025crud}
\bibfield{author}{\bibinfo{person}{Yuanjie Lyu}, \bibinfo{person}{Zhiyu Li}, \bibinfo{person}{Simin Niu}, \bibinfo{person}{Feiyu Xiong}, \bibinfo{person}{Bo Tang}, \bibinfo{person}{Wenjin Wang}, \bibinfo{person}{Hao Wu}, \bibinfo{person}{Huanyong Liu}, \bibinfo{person}{Tong Xu}, {and} \bibinfo{person}{Enhong Chen}.} \bibinfo{year}{2025}\natexlab{}.
\newblock \showarticletitle{Crud-rag: A comprehensive chinese benchmark for retrieval-augmented generation of large language models}.
\newblock \bibinfo{journal}{\emph{ACM Transactions on Information Systems}} \bibinfo{volume}{43}, \bibinfo{number}{2} (\bibinfo{year}{2025}), \bibinfo{pages}{1--32}.
\newblock


\bibitem[Martin et~al\mbox{.}(2024)]%
        {martin2024semantic}
\bibfield{author}{\bibinfo{person}{Andreas Martin}, \bibinfo{person}{Hans~Friedrich Witschel}, \bibinfo{person}{Maximilian Mandl}, {and} \bibinfo{person}{Mona Stockhecke}.} \bibinfo{year}{2024}\natexlab{}.
\newblock \showarticletitle{Semantic Verification in Large Language Model-based Retrieval Augmented Generation}. In \bibinfo{booktitle}{\emph{Proceedings of the AAAI Symposium Series}}, Vol.~\bibinfo{volume}{3}. \bibinfo{pages}{188--192}.
\newblock


\bibitem[Meta(2024)]%
        {llama4}
\bibfield{author}{\bibinfo{person}{Meta}.} \bibinfo{year}{2024}\natexlab{}.
\newblock \bibinfo{title}{LLaMA 4}.
\newblock \bibinfo{howpublished}{\url{https://www.llama.com/models/llama-4}}.
\newblock


\bibitem[Mialon et~al\mbox{.}(2023)]%
        {mialon2023gaia}
\bibfield{author}{\bibinfo{person}{Gr{\'e}goire Mialon}, \bibinfo{person}{Cl{\'e}mentine Fourrier}, \bibinfo{person}{Thomas Wolf}, \bibinfo{person}{Yann LeCun}, {and} \bibinfo{person}{Thomas Scialom}.} \bibinfo{year}{2023}\natexlab{}.
\newblock \showarticletitle{Gaia: a benchmark for general ai assistants}. In \bibinfo{booktitle}{\emph{The Twelfth International Conference on Learning Representations}}.
\newblock


\bibitem[MoonShot\_AI(2025)]%
        {kimi}
\bibfield{author}{\bibinfo{person}{MoonShot\_AI}.} \bibinfo{year}{2025}\natexlab{}.
\newblock \bibinfo{howpublished}{\url{https://kimi.moonshot.cn/}}.
\newblock


\bibitem[Nakano et~al\mbox{.}(2021)]%
        {nakano2021webgpt}
\bibfield{author}{\bibinfo{person}{Reiichiro Nakano}, \bibinfo{person}{Jacob Hilton}, \bibinfo{person}{Suchir Balaji}, \bibinfo{person}{Jeff Wu}, \bibinfo{person}{Long Ouyang}, \bibinfo{person}{Christina Kim}, \bibinfo{person}{Christopher Hesse}, \bibinfo{person}{Shantanu Jain}, \bibinfo{person}{Vineet Kosaraju}, \bibinfo{person}{William Saunders}, {et~al\mbox{.}}} \bibinfo{year}{2021}\natexlab{}.
\newblock \showarticletitle{Webgpt: Browser-assisted question-answering with human feedback}.
\newblock \bibinfo{journal}{\emph{arXiv preprint arXiv:2112.09332}} (\bibinfo{year}{2021}).
\newblock


\bibitem[OpenAI(2024a)]%
        {gpt4o}
\bibfield{author}{\bibinfo{person}{OpenAI}.} \bibinfo{year}{2024}\natexlab{a}.
\newblock \bibinfo{title}{hello-gpt-4o}.
\newblock \bibinfo{howpublished}{\url{https://openai.com/index/hello-gpt-4o/}}.
\newblock


\bibitem[OpenAI(2024b)]%
        {o1}
\bibfield{author}{\bibinfo{person}{OpenAI}.} \bibinfo{year}{2024}\natexlab{b}.
\newblock \bibinfo{title}{Introducing OpenAI o1}.
\newblock \bibinfo{howpublished}{\url{https://openai.com/o1/}}.
\newblock


\bibitem[OpenAI(2025a)]%
        {openai_deep_research}
\bibfield{author}{\bibinfo{person}{OpenAI}.} \bibinfo{year}{2025}\natexlab{a}.
\newblock \bibinfo{title}{Introducing deep research}.
\newblock \bibinfo{howpublished}{\url{https://openai.com/index/introducing-deep-research/}}.
\newblock


\bibitem[OpenAI(2025b)]%
        {o4-mini}
\bibfield{author}{\bibinfo{person}{OpenAI}.} \bibinfo{year}{2025}\natexlab{b}.
\newblock \bibinfo{title}{Introducing OpenAI o3 and o4-mini}.
\newblock \bibinfo{howpublished}{\url{https://openai.com/index/introducing-o3-and-o4-mini/}}.
\newblock


\bibitem[Perplexity(2025)]%
        {perplexity}
\bibfield{author}{\bibinfo{person}{Perplexity}.} \bibinfo{year}{2025}\natexlab{}.
\newblock \bibinfo{howpublished}{\url{https://www.perplexity.ai/}}.
\newblock


\bibitem[Petroni et~al\mbox{.}(2020)]%
        {petroni2020kilt}
\bibfield{author}{\bibinfo{person}{Fabio Petroni}, \bibinfo{person}{Aleksandra Piktus}, \bibinfo{person}{Angela Fan}, \bibinfo{person}{Patrick Lewis}, \bibinfo{person}{Majid Yazdani}, \bibinfo{person}{Nicola De~Cao}, \bibinfo{person}{James Thorne}, \bibinfo{person}{Yacine Jernite}, \bibinfo{person}{Vladimir Karpukhin}, \bibinfo{person}{Jean Maillard}, {et~al\mbox{.}}} \bibinfo{year}{2020}\natexlab{}.
\newblock \showarticletitle{KILT: a benchmark for knowledge intensive language tasks}.
\newblock \bibinfo{journal}{\emph{arXiv preprint arXiv:2009.02252}} (\bibinfo{year}{2020}).
\newblock


\bibitem[Qwen\_Team(2024)]%
        {qwq-32b-preview}
\bibfield{author}{\bibinfo{person}{Qwen\_Team}.} \bibinfo{year}{2024}\natexlab{}.
\newblock \bibinfo{title}{QwQ: Reflect Deeply on the Boundaries of the Unknown}.
\newblock
\newblock
\urldef\tempurl%
\url{https://qwenlm.github.io/blog/qwq-32b-preview/}
\showURL{%
\tempurl}


\bibitem[Qwen\_Team(2025)]%
        {qwen_2_5_max}
\bibfield{author}{\bibinfo{person}{Qwen\_Team}.} \bibinfo{year}{2025}\natexlab{}.
\newblock \bibinfo{title}{Qwen2.5-Max: Exploring the Intelligence of Large-scale MoE Model}.
\newblock \bibinfo{howpublished}{\url{https://qwenlm.github.io/blog/qwen2.5-max/}}.
\newblock


\bibitem[Schick et~al\mbox{.}(2023)]%
        {schick2023toolformer}
\bibfield{author}{\bibinfo{person}{Timo Schick}, \bibinfo{person}{Jane Dwivedi-Yu}, \bibinfo{person}{Roberto Dess{\`\i}}, \bibinfo{person}{Roberta Raileanu}, \bibinfo{person}{Maria Lomeli}, \bibinfo{person}{Eric Hambro}, \bibinfo{person}{Luke Zettlemoyer}, \bibinfo{person}{Nicola Cancedda}, {and} \bibinfo{person}{Thomas Scialom}.} \bibinfo{year}{2023}\natexlab{}.
\newblock \showarticletitle{Toolformer: Language models can teach themselves to use tools}.
\newblock \bibinfo{journal}{\emph{Advances in Neural Information Processing Systems}}  \bibinfo{volume}{36} (\bibinfo{year}{2023}), \bibinfo{pages}{68539--68551}.
\newblock


\bibitem[Tencent(2025)]%
        {yuanbao}
\bibfield{author}{\bibinfo{person}{Tencent}.} \bibinfo{year}{2025}\natexlab{}.
\newblock \bibinfo{howpublished}{\url{https://yuanbao.tencent.com/chat/}}.
\newblock


\bibitem[Thorne et~al\mbox{.}(2018)]%
        {thorne2018fever}
\bibfield{author}{\bibinfo{person}{James Thorne}, \bibinfo{person}{Andreas Vlachos}, \bibinfo{person}{Christos Christodoulopoulos}, {and} \bibinfo{person}{Arpit Mittal}.} \bibinfo{year}{2018}\natexlab{}.
\newblock \showarticletitle{FEVER: a large-scale dataset for fact extraction and VERification}.
\newblock \bibinfo{journal}{\emph{arXiv preprint arXiv:1803.05355}} (\bibinfo{year}{2018}).
\newblock


\bibitem[Vu et~al\mbox{.}(2023)]%
        {vu2023freshllms}
\bibfield{author}{\bibinfo{person}{Tu Vu}, \bibinfo{person}{Mohit Iyyer}, \bibinfo{person}{Xuezhi Wang}, \bibinfo{person}{Noah Constant}, \bibinfo{person}{Jerry Wei}, \bibinfo{person}{Jason Wei}, \bibinfo{person}{Chris Tar}, \bibinfo{person}{Yun-Hsuan Sung}, \bibinfo{person}{Denny Zhou}, \bibinfo{person}{Quoc Le}, {et~al\mbox{.}}} \bibinfo{year}{2023}\natexlab{}.
\newblock \showarticletitle{Freshllms: Refreshing large language models with search engine augmentation}.
\newblock \bibinfo{journal}{\emph{arXiv preprint arXiv:2310.03214}} (\bibinfo{year}{2023}).
\newblock


\bibitem[Wei et~al\mbox{.}(2025)]%
        {wei2025browsecomp}
\bibfield{author}{\bibinfo{person}{Jason Wei}, \bibinfo{person}{Zhiqing Sun}, \bibinfo{person}{Spencer Papay}, \bibinfo{person}{Scott McKinney}, \bibinfo{person}{Jeffrey Han}, \bibinfo{person}{Isa Fulford}, \bibinfo{person}{Hyung~Won Chung}, \bibinfo{person}{Alex~Tachard Passos}, \bibinfo{person}{William Fedus}, {and} \bibinfo{person}{Amelia Glaese}.} \bibinfo{year}{2025}\natexlab{}.
\newblock \showarticletitle{BrowseComp: A Simple Yet Challenging Benchmark for Browsing Agents}.
\newblock \bibinfo{journal}{\emph{arXiv preprint arXiv:2504.12516}} (\bibinfo{year}{2025}).
\newblock


\bibitem[Wiratunga et~al\mbox{.}(2024)]%
        {wiratunga2024cbr}
\bibfield{author}{\bibinfo{person}{Nirmalie Wiratunga}, \bibinfo{person}{Ramitha Abeyratne}, \bibinfo{person}{Lasal Jayawardena}, \bibinfo{person}{Kyle Martin}, \bibinfo{person}{Stewart Massie}, \bibinfo{person}{Ikechukwu Nkisi-Orji}, \bibinfo{person}{Ruvan Weerasinghe}, \bibinfo{person}{Anne Liret}, {and} \bibinfo{person}{Bruno Fleisch}.} \bibinfo{year}{2024}\natexlab{}.
\newblock \showarticletitle{CBR-RAG: case-based reasoning for retrieval augmented generation in LLMs for legal question answering}. In \bibinfo{booktitle}{\emph{International Conference on Case-Based Reasoning}}. Springer, \bibinfo{pages}{445--460}.
\newblock


\bibitem[xAI(2025)]%
        {grok3}
\bibfield{author}{\bibinfo{person}{xAI}.} \bibinfo{year}{2025}\natexlab{}.
\newblock \bibinfo{howpublished}{\url{https://grok.com/}}.
\newblock


\bibitem[Xi et~al\mbox{.}(2025)]%
        {xi2025rise}
\bibfield{author}{\bibinfo{person}{Zhiheng Xi}, \bibinfo{person}{Wenxiang Chen}, \bibinfo{person}{Xin Guo}, \bibinfo{person}{Wei He}, \bibinfo{person}{Yiwen Ding}, \bibinfo{person}{Boyang Hong}, \bibinfo{person}{Ming Zhang}, \bibinfo{person}{Junzhe Wang}, \bibinfo{person}{Senjie Jin}, \bibinfo{person}{Enyu Zhou}, {et~al\mbox{.}}} \bibinfo{year}{2025}\natexlab{}.
\newblock \showarticletitle{The rise and potential of large language model based agents: A survey}.
\newblock \bibinfo{journal}{\emph{Science China Information Sciences}} \bibinfo{volume}{68}, \bibinfo{number}{2} (\bibinfo{year}{2025}), \bibinfo{pages}{121101}.
\newblock


\bibitem[Xiong et~al\mbox{.}(2024)]%
        {xiong2024search}
\bibfield{author}{\bibinfo{person}{Haoyi Xiong}, \bibinfo{person}{Jiang Bian}, \bibinfo{person}{Yuchen Li}, \bibinfo{person}{Xuhong Li}, \bibinfo{person}{Mengnan Du}, \bibinfo{person}{Shuaiqiang Wang}, \bibinfo{person}{Dawei Yin}, {and} \bibinfo{person}{Sumi Helal}.} \bibinfo{year}{2024}\natexlab{}.
\newblock \showarticletitle{When search engine services meet large language models: visions and challenges}.
\newblock \bibinfo{journal}{\emph{IEEE Transactions on Services Computing}} (\bibinfo{year}{2024}).
\newblock


\bibitem[Xu et~al\mbox{.}(2024)]%
        {xu2024let}
\bibfield{author}{\bibinfo{person}{Zhikun Xu}, \bibinfo{person}{Yinghui Li}, \bibinfo{person}{Ruixue Ding}, \bibinfo{person}{Xinyu Wang}, \bibinfo{person}{Boli Chen}, \bibinfo{person}{Yong Jiang}, \bibinfo{person}{Hai-Tao Zheng}, \bibinfo{person}{Wenlian Lu}, \bibinfo{person}{Pengjun Xie}, {and} \bibinfo{person}{Fei Huang}.} \bibinfo{year}{2024}\natexlab{}.
\newblock \showarticletitle{Let llms take on the latest challenges! a chinese dynamic question answering benchmark}.
\newblock \bibinfo{journal}{\emph{arXiv preprint arXiv:2402.19248}} (\bibinfo{year}{2024}).
\newblock


\bibitem[Yang et~al\mbox{.}(2024)]%
        {yang2024qwen2}
\bibfield{author}{\bibinfo{person}{An Yang}, \bibinfo{person}{Baosong Yang}, \bibinfo{person}{Beichen Zhang}, \bibinfo{person}{Binyuan Hui}, \bibinfo{person}{Bo Zheng}, \bibinfo{person}{Bowen Yu}, \bibinfo{person}{Chengyuan Li}, \bibinfo{person}{Dayiheng Liu}, \bibinfo{person}{Fei Huang}, \bibinfo{person}{Haoran Wei}, {et~al\mbox{.}}} \bibinfo{year}{2024}\natexlab{}.
\newblock \showarticletitle{Qwen2. 5 technical report}.
\newblock \bibinfo{journal}{\emph{arXiv preprint arXiv:2412.15115}} (\bibinfo{year}{2024}).
\newblock


\bibitem[Yang et~al\mbox{.}(2018)]%
        {yang2018hotpotqa}
\bibfield{author}{\bibinfo{person}{Zhilin Yang}, \bibinfo{person}{Peng Qi}, \bibinfo{person}{Saizheng Zhang}, \bibinfo{person}{Yoshua Bengio}, \bibinfo{person}{William~W Cohen}, \bibinfo{person}{Ruslan Salakhutdinov}, {and} \bibinfo{person}{Christopher~D Manning}.} \bibinfo{year}{2018}\natexlab{}.
\newblock \showarticletitle{HotpotQA: A dataset for diverse, explainable multi-hop question answering}.
\newblock \bibinfo{journal}{\emph{arXiv preprint arXiv:1809.09600}} (\bibinfo{year}{2018}).
\newblock


\bibitem[Yao et~al\mbox{.}(2023)]%
        {yao2023react}
\bibfield{author}{\bibinfo{person}{Shunyu Yao}, \bibinfo{person}{Jeffrey Zhao}, \bibinfo{person}{Dian Yu}, \bibinfo{person}{Nan Du}, \bibinfo{person}{Izhak Shafran}, \bibinfo{person}{Karthik Narasimhan}, {and} \bibinfo{person}{Yuan Cao}.} \bibinfo{year}{2023}\natexlab{}.
\newblock \showarticletitle{React: Synergizing reasoning and acting in language models}. In \bibinfo{booktitle}{\emph{International Conference on Learning Representations (ICLR)}}.
\newblock


\end{thebibliography}

%%%%%%%%%%%%%%%%%%%%%%%%%%%%%%%%%%%%%%%%%%%%%%%%%%%%%%%%%%%%

\newpage
\appendix

\section{Technical Appendices and Supplementary Material}
\subsection{Instructions for model prediction}
In the evaluation, we followed the BrowseComp instructions for model predictions. Additionally, for both open-source and closed-source models, which may exhibit a refusal to answer (often stating the lack of search capabilities), we included an instruction that prompts the models to rely on their intrinsic knowledge for providing answers.

\begin{figure}[h]
  \centering
  \includegraphics[width=1\textwidth]{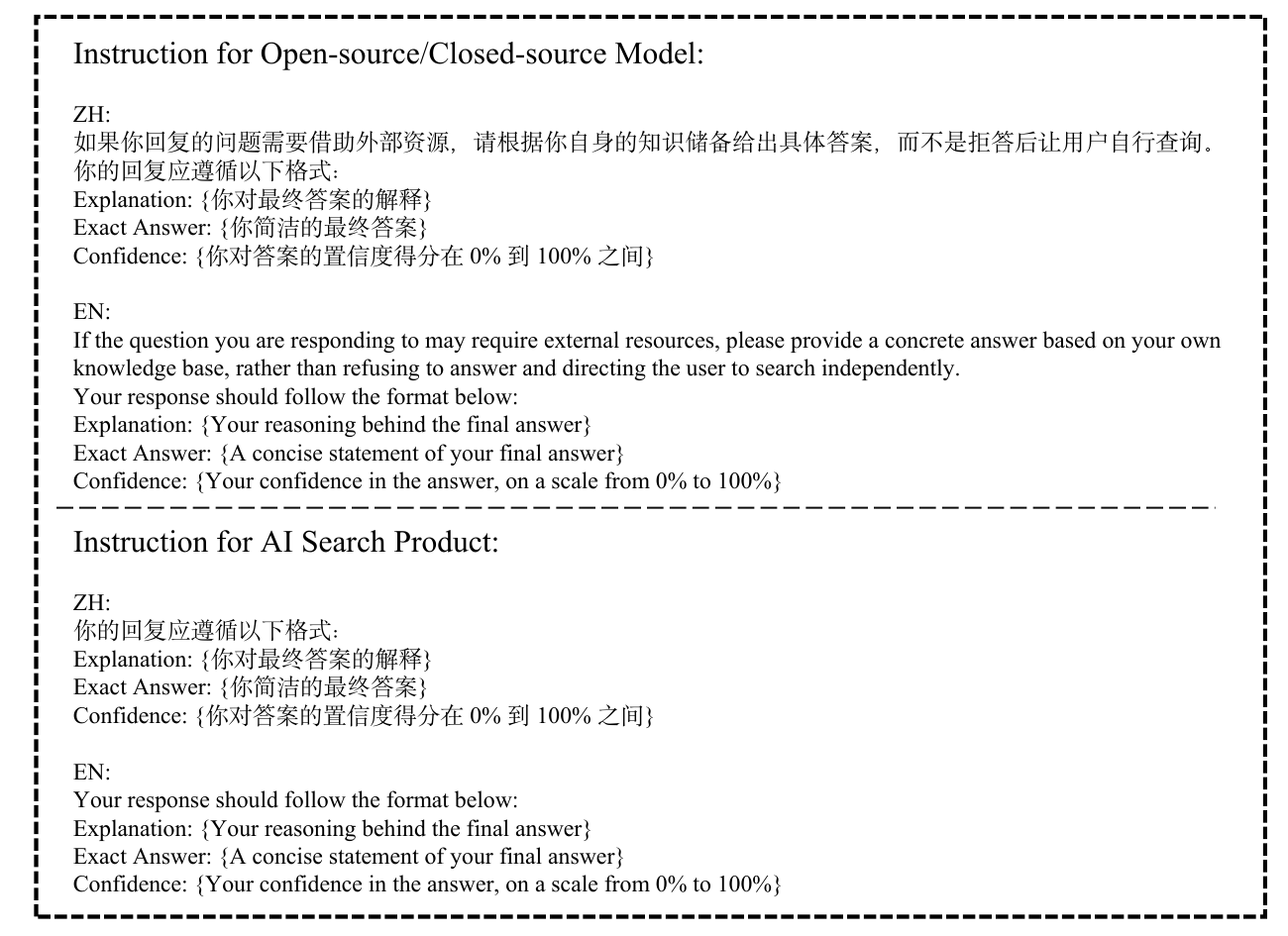}
  \caption{Instruction for model prediction.}
  \label{fig:instruction}
\end{figure}

\subsection{Instruction for grading}
We adopt the same grading prompt as used in BrowseComp and employ GPT-4o for the grading process.

\begin{figure}[h]
  \centering
  \includegraphics[width=1\textwidth]{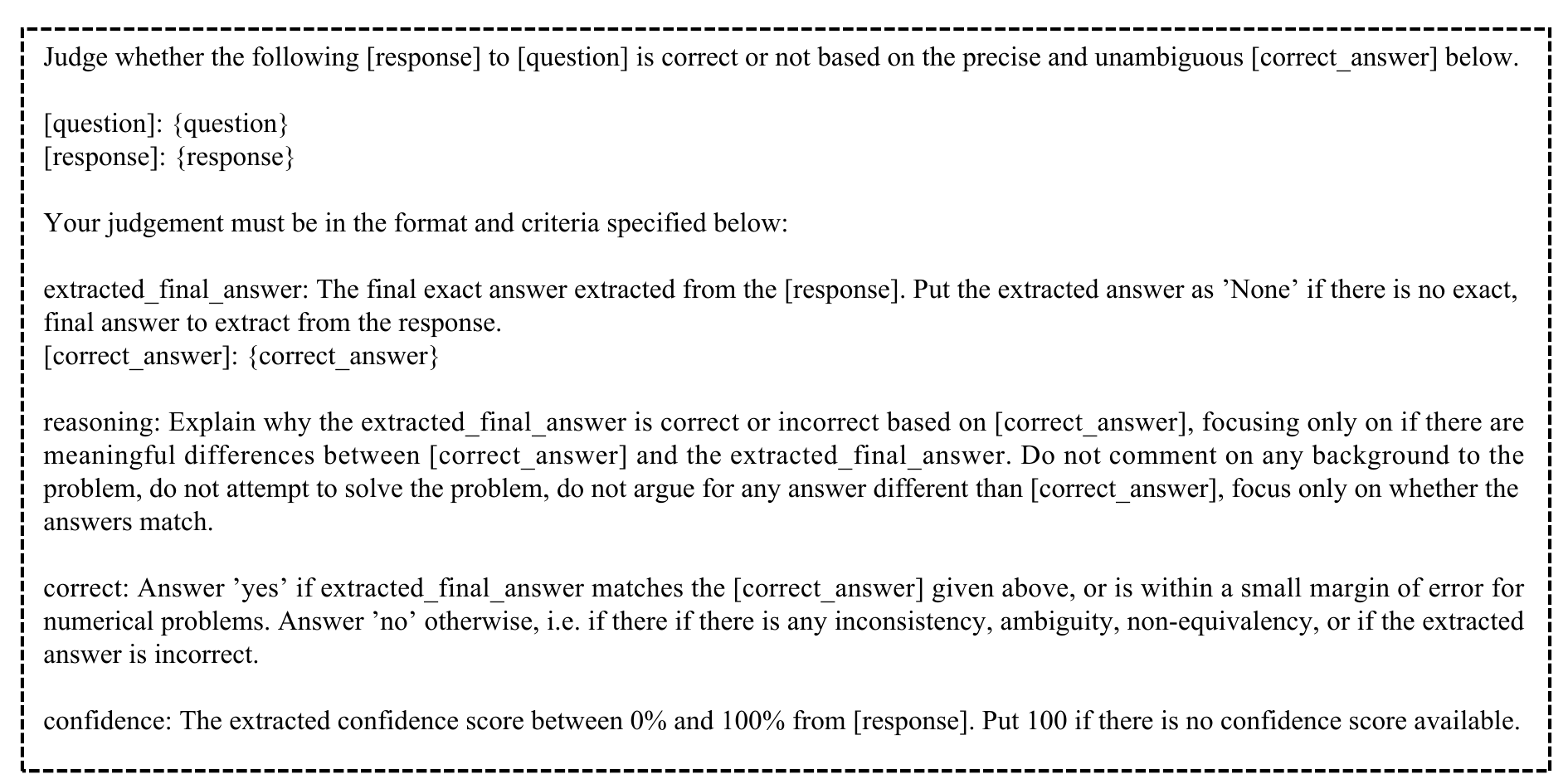}
  \caption{Prompt for model grading.}
  \label{fig:judge_prompt}
\end{figure}

\end{document}